\documentclass[letterpaper, 10 pt, conference]{ieeeconf}  % Comment this line out if you need a4paper

\IEEEoverridecommandlockouts                              % This command is only needed if 
\usepackage{graphicx}
\usepackage{makecell}
\usepackage{multirow}
\usepackage{amsmath}
\usepackage{subfigure}
\usepackage{color}
\usepackage{caption}
\usepackage{amssymb}
\usepackage{stfloats}
\usepackage{hyperref}
\title{\LARGE \bf
DAM-VLA: A Dynamic Action Model-Based Vision-Language-Action Framework for Robot Manipulation
}

\author{Xiongfeng Peng$^{1}$, Jiaqian Yu$^{1}$, Dingzhe Li$^{1}$, Yixiang Jin$^{1}$, Lu Xu$^{1}$, Yamin Mao$^{1}$, Chao Zhang$^{1}$, \\ Weiming Li$^{1}$, Sujin Jang$^{2,3}$, Dongwook Lee$^{2}$, and Daehyun Ji$^{2}$% <-this % stops a space
\thanks{$^{1}$Xiongfeng Peng, Jiaqian Yu, Dingzhe Li, Yixiang Jin, Lu Xu, Yamin Mao, Chao Zhang, and Weiming Li are with Advanced Research Lab, Samsung R\&D Institute China-Beijing (SRCB), China}%
\thanks{$^{2}$Sujin Jang, Dongwook Lee, and Daehyun Ji are with Samsung AI Center, DS Division, South Korea}%
\thanks{$^{3}$Sujin Jang is also with Hanyang University ERICA, South Korea}%
}

\begin{document}

%%% Teaser Fig. Option #1
\IEEEaftertitletext{
  \centering
  % \vspace{-20pt} % Adjust spacing as needed
  \includegraphics[width=1.0\textwidth,height=5cm]{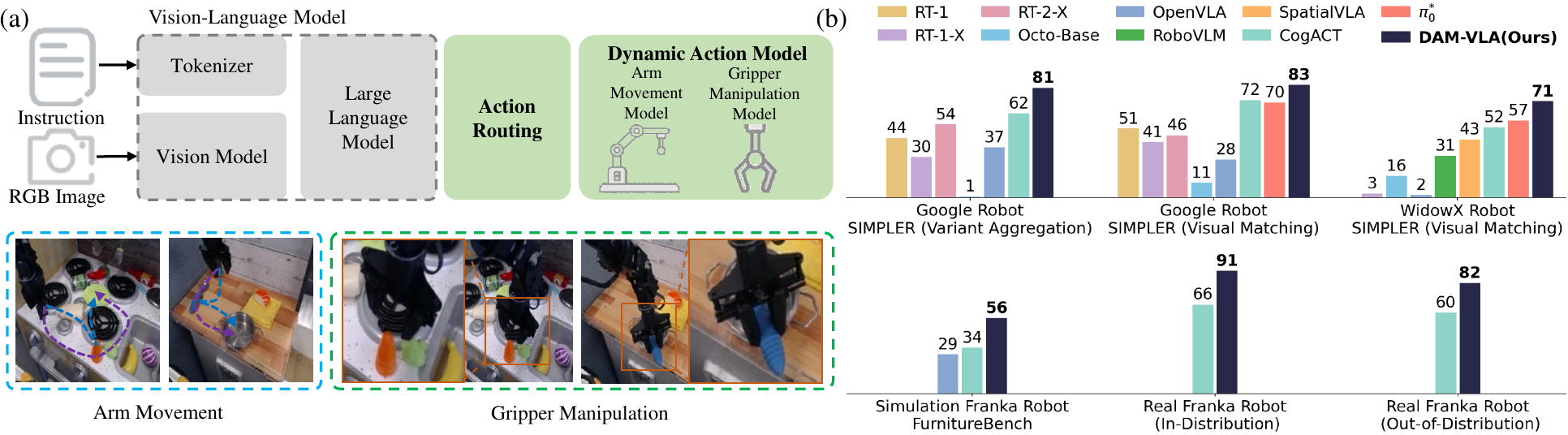}
  \captionof{figure}{DAM-VLA framework and experimental results. (a) We propose a DAM-VLA framework that dynamically integrates the inherent reasoning capabilities of VLMs with specialized diffusion-based action models tailored for arm movement and gripper manipulation. In various robotic tasks, arm movement typically covers a larger spatial range than gripper manipulation. consequently, in the observed images, the trajectories of the arm movement often occupy the majority of the region, while gripper manipulation is usually confined to a small, localized area; (b) Across extensive evaluations, our DAM-VLA achieves superior average success rates compared to state-of-the-art VLA methods, demonstrating improvements in both pick-and-place tasks within the SIMPLER simulation and long-horizon tasks on the FurnitureBench simulation, as well as in real-world pick-and-place evaluations.} 
  \label{motivation}
  \vspace{10pt}
}
\maketitle
\thispagestyle{empty}
\pagestyle{empty}

%%%%%%%%%%%%%%%%%%%%%%%%%%%%%%%%%%%%%%%%%%%%%%%%%%%%%%%%%%%%%%%%%%%%%%%%%%%%%%%%
\begin{abstract}

In dynamic environments such as warehouses, hospitals, and homes, robots must seamlessly transition between gross motion and precise manipulations to complete complex tasks. However, current Vision-Language-Action (VLA) frameworks, largely adapted from pre-trained Vision-Language Models (VLMs), often struggle to reconcile general task adaptability with the specialized precision required for intricate manipulation. To address this challenge, we propose DAM-VLA, a dynamic action model-based VLA framework. DAM-VLA integrates VLM reasoning with diffusion-based action models specialized for arm and gripper control. Specifically, it introduces (i) an action routing mechanism, using task-specific visual and linguistic cues to select appropriate action models (e.g., arm movement or gripper manipulation), (ii) a dynamic action model that fuses high-level VLM cognition with low-level visual features to predict actions, and (iii) a dual-scale action weighting mechanism that enables dynamic coordination between the arm-movement and gripper-manipulation models. Across extensive evaluations, DAM-VLA achieves superior success rates compared to state-of-the-art VLA methods in simulated (SIMPLER, FurnitureBench) and real-world settings, showing robust generalization from standard pick-and-place to demanding long-horizon and contact-rich tasks.
\end{abstract}

\section{Introduction}
% Method:
% DP, DP3， ACT
% DiT Policy, octo, HPT
% OpenVLA, RT-2
% RT-H, ECOT, Diffusion-VLA, CoA-VLA, RT-Affordance
% TinyVLA, RDT-1B, RoboDual, CogAct, HybridVLA, RoboMamba, Dex-VLA
% RoboMatrix

% Summary:
% SIMPLER
% VLATest(An Empirical Study)

% ACT, Diffusion Policy
A central challenge in robotics is enabling robots to perform diverse tasks in dynamic environments. Conventional robot learning methods typically train policies on datasets curated for a specific robot and task. The resulting policies act as specialists, such as the popular ACT \cite{zhao2023learning} and Diffusion Policy \cite{chi2023diffusion}. Although these approaches achieve high precision in targeted scenarios, they generalize poorly across varying environments and tasks.

\begin{figure*}[h]
\setlength{\abovecaptionskip}{0.2cm}
\setlength{\belowcaptionskip}{-0.4cm}
    \centering
    \includegraphics[width=1.0\linewidth]{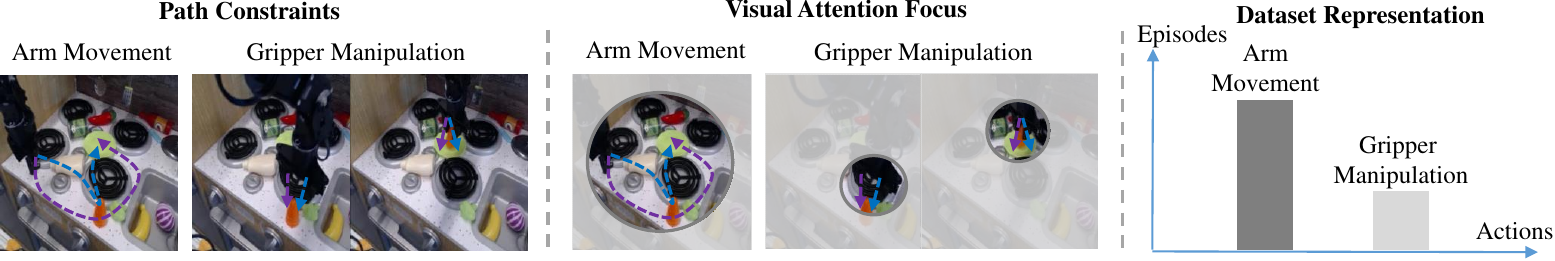}
    \captionsetup{justification=justified, singlelinecheck=false}
    \caption{We identify three distinctions between the arm movement and the gripper manipulation using the task of placing a carrot on a plate as an illustrative example: Path Constrains, Visual Attention, and Dataset Representation.}
    \label{motivitation}
    % \vspace{-1.2em}
    \end{figure*}

% OpenVLA, RT-2
Recently, VLA models have attracted attention for their ability to extend pretrained VLMs to robotics by discretizing continuous actions into bins for action prediction. Representative works such as RT-2 \cite{brohan2023rt} and OpenVLA \cite{kim2024openvla} have demonstrated impressive performance in multi-task learning and generalization. By enabling robots to interpret visual observations and language instructions, VLA models can generate generalizable action sequences. Consequently, leveraging the inherent capabilities of VLMs in developing VLA frameworks is crucial for achieving both task-specific precision and broad generalization in dynamic environments.

On the one hand, several existing VLA methods, such as RT-H \cite{belkhale2024rt}, RT-Affordance \cite{nasiriany2024rt}, and ECOT \cite{zawalski2024robotic}, incorporate Chain-of-Thought (CoT) reasoning to analyze spatial relationships between the gripper and the object, and predict the corresponding actions. Although CoT enhances the reasoning capabilities of VLMs and improves generalization, it introduces many extra reasoning tokens and substantially increase inference time. 
On the other hand, diffusion-based VLA methods, including $\pi0$ \cite{black2410pi0}, TinyVLA \cite{wen2025tinyvla}, RDT-1B \cite{liu2024rdt}, RoboDual \cite{bu2024towards}, CogACT \cite{li2024cogact}, and HybridVLA \cite{liu2025hybridvla}, append a separate diffusion head after the VLM. These methods either condition on VLM-extracted features or jointly embed the denoising timestep and noisy actions into the token sequence during diffusion. While enabling more precise manipulation, relying solely on VLM-extracted features limits the integration of richer multi-modal cues.

To better leverage the inherent capabilities of VLMs for VLA models that combine both task-specific and general manipulation in dynamic environments, we first identify several key distinctions between arm movement and gripper manipulation.
In many robotic tasks, arm movement spans a larger spatial range than gripper manipulation. Consequently, arm trajectories often dominate the scene in the observed images, while gripper manipulations are confined to small, localized regions. Figure 2 illustrates this disparity using the task of placing a carrot on a plate as an example, where the non-gray regions of visual attention highlight the difference. Specifically, arm movement requires global attention, whereas gripper manipulation demands localized focus.
The fundamental distinctions can be summarized as follows:
(1) \textbf{Path Constraints}. Arm movement is relatively unconstrained since the robot can take multiple trajectories to reach the carrot. In contrast, gripper manipulation is highly constrained, requiring precise grasping postures for success.
(2) \textbf{Visual Attention}. Arm movement depends on global scene understanding, whereas gripper manipulation necessitates fine-grained, localized visual attention.
(3) \textbf{Dataset Representation}. Datasets usually contain far more arm movement episodes than gripper manipulation ones. Nevertheless, despite being fewer, gripper manipulations are critical for task success and often more complex.

Building on these distinctions, we leverage VLM reasoning to differentiate action types (arm movement vs. gripper manipulation), and apply the corresponding action model to perform the required manipulation. Rather than loosely coupling a VLM with separate action models, we introduce the \textbf{DAM-VLA} framework (Figure 1), which fully exploits the strengths of VLMs to support both task-specific precision and generalization in dynamic environments. The main contributions of this work are summarized as follows:
(1) \noindent\textbf{Action Routing.} A VLM-guided router interprets task-specific visual and linguistic cues to select the appropriate action models (e.g., arm movement or gripper manipulation).
(2) \noindent\textbf{Dynamic Action Model.} A dual-head diffusion model that integrates high-level cognition from the VLM with low-level visual information to predict actions across different models.
(3) \noindent\textbf{Dual-Scale Action Weighting.} A two-scale weighting mechanism (i.e., trajectory-level and action-chunk-level) enables dynamic coordination between the arm-movement and gripper-manipulation models.
(4) \noindent\textbf{Extensive Evaluation.} DAM-VLA achieves superior average success rates compared to state-of-the-art VLA methods, across both pick-and-place tasks in the SIMPLER simulation \cite{li2024evaluating} and long-horizon tasks in the FurnitureBench simulation \cite{heo2023furniturebench}, as well as in real-world pick-and-place experiments.

%===============================================================================

\section{Related Work}
\label{sec:related-work}
\noindent\textbf{Vision-Language-Action Models.}
%\subsection{Vision-Language-Action Models}
LLMs \cite{achiam2023gpt, touvron2023llama, touvron2023llama2} and VLMs \cite{alayrac2022flamingo, team2023gemini, bai2023qwen, li2023blip, liu2023visual} inspire the development of VLA models, which extend VLMs by integrating action generation. RT-2 \cite{brohan2023rt} tokenizes 7D actions into discrete tokens and employs the VLM PaLI-X \cite{chen2023pali} for prediction, while OpenVLA \cite{kim2024openvla} follows a similar approach with the Prismatic VLM \cite{karamcheti2024prismatic}. RT-H \cite{belkhale2024rt}, RT-Affordance \cite{nasiriany2024rt}, and ECoT \cite{zawalski2024robotic} incorporate Chain-of-Thought (CoT) reasoning to analyze spatial relationships between the gripper and objects before predicting actions.
While these methods more effectively exploit the reasoning capabilities of pretrained VLMs to improve generalization, the large number of additional reasoning tokens (e.g., 7 in OpenVLA vs. 350 in ECoT) substantially slows inference and reduces control frequency. Moreover, they do not adequately address the continuity, accuracy, and specificity required for precise action estimation.

\begin{figure*}[h]
\setlength{\abovecaptionskip}{0.2cm}
\setlength{\belowcaptionskip}{-0.4cm}
    \centering
    \includegraphics[width=0.95\linewidth]{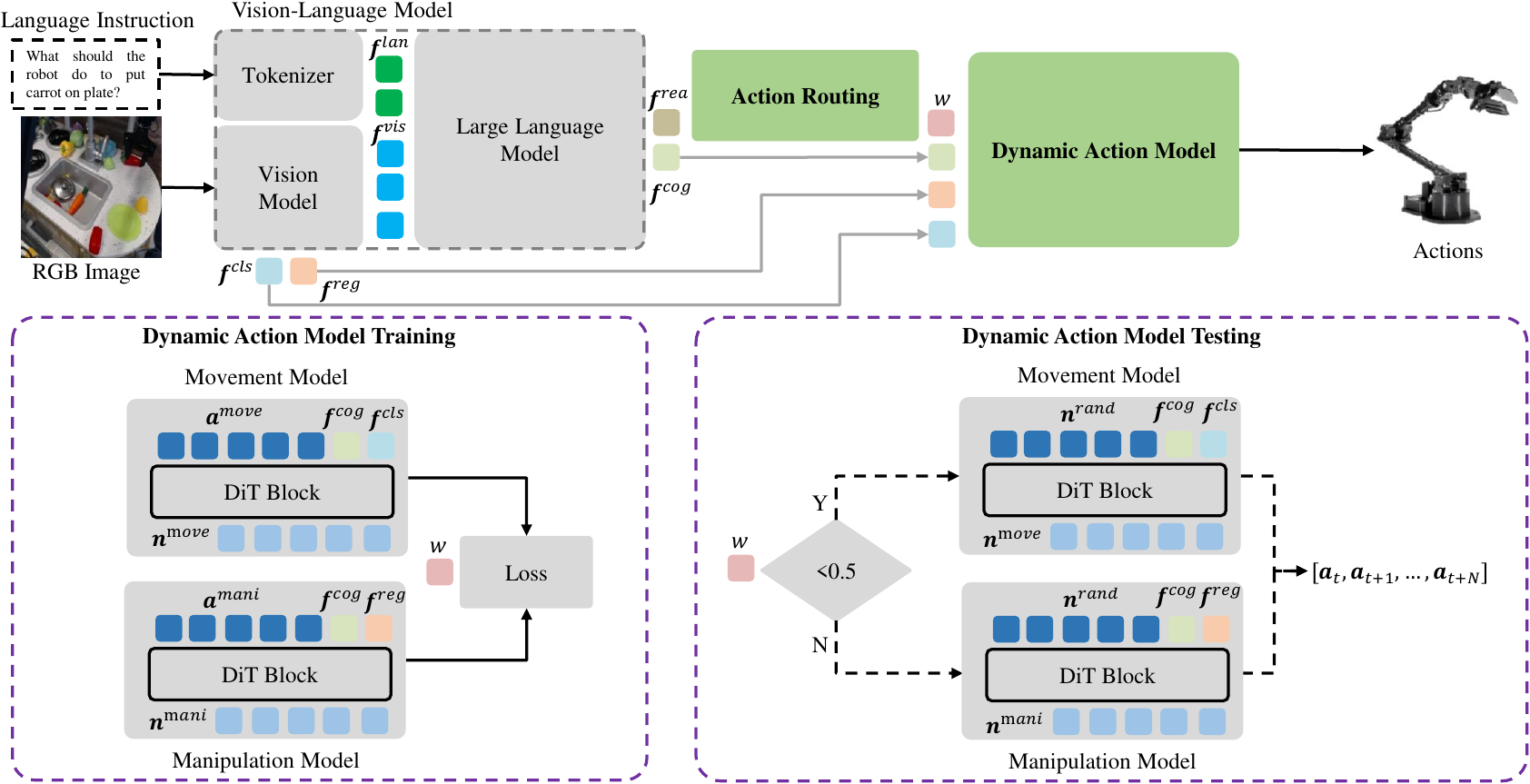}
    \captionsetup{justification=justified, singlelinecheck=false}
    \caption{The architecture of our DAM-VLA. Given an RGB image observation and a task description, the model predicts a sequence of temporal actions. The process consists of three key components: 1) a vision-language model that encodes observation into visual, class and register tokens, and integrates visual tokens with a set of linguistic tokens, and produces the cognition and reasoning latents; 2) an action routing module that generates a weight and feeds it into the dynamic action model; 3) a dynamic action model that dynamically executes different action models by combining the low-level class token or register token from the vision model with the high-level cognition latent from the VLM to predict an action sequence.
    }
    \label{motivitation}
    % \vspace{-1.0em}
    \end{figure*}

\noindent\textbf{Diffusion Action Models.}
%\subsection{Diffusion Action Models}
The Diffusion Policy \cite{chi2023diffusion} applies diffusion models \cite{ho2020denoising, peebles2023scalable} to robot learning, demonstrating the ability to model multimodal action distributions. Subsequent research \cite{ho2020denoising, peebles2023scalable, black2023zero, dasari2024ingredients, ren2024diffusion, reuss2024multimodal, uehara2024fine, uehara2024feedback, zhao2024aloha, ke20243d, jia2024mail, team2024octo} has further advanced this line of work by extending diffusion policies to 3D environments, scaling their capabilities, improving efficiency, and introducing architectural innovations. Among these, Octo \cite{team2024octo} augments a transformer-based backbone with compact diffusion heads, enabling adaptation of action outputs across different robots.
Although diffusion-based models improve efficiency and performance across diverse tasks and robotic platforms, they have yet to exploit pretrained LLMs and VLMs, whose strong generalization and reasoning capabilities could further enhance policy robustness.

\noindent\textbf{Diffusion-based Vision-Language-Action Models.}
%\subsection{Diffusion-Based Vision-Language-Action Models}
To combine the strengths of VLMs and diffusion-based action models, $\pi0$ \cite{black2410pi0} introduces a separate diffusion head to generate actions via flow matching, while TinyVLA \cite{wen2025tinyvla} attaches a lightweight diffusion head after a compact VLM. CogACT \cite{li2024cogact} and DiVLA \cite{wen2024diffusion} decouple reasoning and action prediction, assigning these functions to the VLM and the diffusion head, respectively. HybridVLA \cite{liu2025hybridvla} further integrates diffusion and autoregressive action prediction within a single LLM.
However, these approaches do not fully exploit the inherent capabilities of VLMs to build VLA models that achieve both task-specific precision and generalizable manipulation in dynamic environments.

%===============================================================================

\section{Method}
\label{sec:method}
In this section, we first describe the overall DAM-VLA architecture in Section~\ref{sec:archi}. Then we introduce the VLM in Section~\ref{sec:vlm}. To fully leverage the specific manipulation capabilities of different action models and the VLM’s inherent reasoning capabilities, we introduce an action routing mechanism and our dynamic action model in Section~\ref{sec:action_routing}. To further enhance robustness, we propose a dual-scale action weighting mechanism in Section~\ref{sec:action_weighting}.

\subsection{Overall Architecture}\label{sec:archi}

Our goal is to develop a dynamic action model-based VLA framework that enables different robots to physically execute diverse tasks in dynamic environments while receiving an RGB image observation and a task description in the form of a language instruction. Formally, given the language instruction $\boldsymbol{l}$ and visual observation $\boldsymbol{o}_t$ at time $t$, the model $\pi$ predicts a temporal action sequence $[\boldsymbol{a}_{t}, \boldsymbol{a}_{t+1}, ..., \boldsymbol{a}_{t+N}] = \pi(\boldsymbol{l}, \boldsymbol{o}_t)$. The action space $\boldsymbol{a}_{t} = [\delta{\boldsymbol{x}}, \delta{\boldsymbol{\theta}}, s^{grip}]$ corresponds to the gripper with 7 degrees of freedom (DoF), where $\delta{\boldsymbol{x}}$ represents the relative translation offsets of the end effector, $\delta{\boldsymbol{\theta}}$ denotes the rotational changes, and $s^{grip} \in \{0, 1\}$ indicates the gripper’s open or close state.

In Figure 3, the architecture of DAM-VLA is shown to consist of three key components:
1) A vision-language model, that encodes information from observation $\boldsymbol{o}_t$ into visual tokens $\boldsymbol{f}^{vis}$, a class token $\boldsymbol{f}^{cls}$, and a register token $\boldsymbol{f}^{reg}$, and integrates visual tokens $\boldsymbol{f}^{vis}$ with a set of linguistic tokens $\boldsymbol{f}^{lan}$ from a language instruction $\boldsymbol{l}$, and produces the cognition latent $\boldsymbol{f}^{cog}$ and reasoning latent $\boldsymbol{f}^{rea}$;
2) An action routing module that generates a weight $w$ and feeds it into the dynamic action model;
3) A dynamic action model that dynamically executes different action models by combining the low-level class token $\boldsymbol{f}^{cls}$ or register token $\boldsymbol{f}^{reg}$ from the vision model with the high-level cognition latent $\boldsymbol{f}^{cog}$ from the VLM to predict an action sequence $[\boldsymbol{a}_{t}, \boldsymbol{a}_{t+1}, ..., \boldsymbol{a}_{t+N}]$.

\subsection{Vision-Language Model}\label{sec:vlm}
The vision model processes the RGB image input into a set of tokens, which include not only visual tokens $\boldsymbol{f}^{vis}$, but also a class token $\boldsymbol{f}^{cls}$ and a register token $\boldsymbol{f}^{reg}$ from DINOv2 \cite{oquab2023dinov2}. The vision model consists of powerful vision transformers, DINOv2 and SigLIP \cite{zhai2023sigmoid}, pretrained on internet-scale image data to capture both low-level rich visual features and high-level semantic understanding.
At each timestep $t$, the image observation $\boldsymbol{o}_t$ is fed into both the arm movement model and the gripper manipulation model, each producing a downsampled feature map. These feature maps are then concatenated along the channel dimension, passed through a linear projection layer, and serialized into a set of tokens, including visual tokens $\boldsymbol{f}^{vis} = [\boldsymbol{v}_1, \boldsymbol{v}_2, ..., \boldsymbol{v}_{NV}]$, a class token $\boldsymbol{f}^{cls}$, and a register token $\boldsymbol{f}^{reg}$.

The large language model is responsible for integrating visual information and language instructions and estimating the reasoning token. We use an LLaMA-2 model \cite{touvron2023llama2} as the backbone. The language instruction $\boldsymbol{l}$ is tokenized into a set of linguistic tokens, $\boldsymbol{f}^{lan} = [\boldsymbol{l}_1, \boldsymbol{l}_2, ..., \boldsymbol{l}_{NL}]$. These tokens $\boldsymbol{f}^{lan}$ are then concatenated with the visual tokens $\boldsymbol{f}^{vis}$ and an additional learnable reasoning token, and are processed by the large language model using a causal attention mechanism.
The resulting output consists of the cognition and reasoning latents, $\boldsymbol{f}^{cog}$ and $\boldsymbol{f}^{rea}$, respectively. $\boldsymbol{f}^{rea}$ and $\boldsymbol{f}^{cog}$ are derived from the hidden features of the second and last transformer layers of the LLM, respectively. $\boldsymbol{f}^{rea}$ serves as the input to the subsequent action routing module to select the appropriate action models, while $\boldsymbol{f}^{cog}$ serves as the input for the action model to predict the actions.

\subsection{Action Routing Mechanism and Dynamic Action Model}\label{sec:action_routing}
% \subsubsection{Action Routing Mechanism}
To determine whether the action state is in arm movement or gripper manipulation, we design an action routing mechanism. It leverages the reasoning ability of the VLM to learn the weight $w$ and it is supervised by the labeled weight $\hat{w} \in {0, 1}$ we designed for dynamically executing either the arm movement model or the gripper manipulation model.
The labeled weight $\hat{w}$ is obtained based on the state of the robot gripper. When the state of the robot gripper changes from open to closed or vice versa, we define this as a transition of the robot action from arm movement ($\hat{w}=0$) to gripper manipulation ($\hat{w}=1$). The detailed calculation process for $\hat{w}$ is explained in the a dual-scale action weighting section.
To leverage the reasoning ability of the VLM, the reasoning latent $\boldsymbol{f}^{rea}$ from the VLM is input to the action routing module, where it is processed through a fully connected layer and a normalization layer. The output is the predicted weight $w$, which is supervised by the following cross-entropy loss:
\begin{equation}\label{equ:cross_entropy_loss}
\begin{aligned}
 L_{class} = ||-(\hat{w}\log(w)+(1-\hat{w})\log(1-w))||^{1}.
\end{aligned}
\end{equation}

To fully leverage the specific manipulation capabilities of different diffusion action models and the VLM’s inherent reasoning capabilities, we propose the dynamic action model. %Next, we will introduce the training and testing processes in detail.
In the training phase, the dynamic action model enables targeted learning of the arm movement model and the gripper manipulation model using the labeled weights $\boldsymbol{\hat{w}}^{move}$ and $\boldsymbol{\hat{w}}^{mani}$. These weights are calculated by the trajectory weights and the action chunk weights. A more detailed explanation of the design of $\boldsymbol{\hat{w}}^{move}$ and $\boldsymbol{\hat{w}}^{mani}$ is provided in the dual-scale action weighting section.
In addition to the high-level VLM-extracted cognition latent $\boldsymbol{f}^{cog}$ as a condition, the two action models also take lower-level visual tokens $\boldsymbol{f}^{cls}$ or $\boldsymbol{f}^{reg}$ as conditions, respectively. Since the two models focus on different attention focuses in the image, we input different visual tokens. The arm movement model requires more global attention, so we input the class token $\boldsymbol{f}^{cls}$ as a condition. In contrast, the gripper manipulation model focuses on more local attention, so we input the register token $\boldsymbol{f}^{reg}$.
Regarding the Diffusion Transformer (DiT) \cite{peebles2023scalable} block of the action models, we refer to \cite{li2024cogact}. The loss functions for the arm movement and the gripper manipulation models are supervised separately as follows:
\begin{equation}\label{equ:movement_loss}
\begin{aligned}
 L_{move} = ||\boldsymbol{n}^{move}_{i} - \boldsymbol{\hat{n}}^{move}||^{2}_{\sum{\boldsymbol{\hat{w}}^{move}}},
\end{aligned}
\end{equation}
\begin{equation}\label{equ:manipulation_loss}
\begin{aligned}
 L_{mani} = ||\boldsymbol{n}^{mani}_{i} - \boldsymbol{\hat{n}}^{mani}||^{2}_{\sum{\boldsymbol{\hat{w}}^{mani}}},
\end{aligned}
\end{equation}
where $||\cdot||_{\sum}$ denotes the Mahalanobis distance, which weights the error terms based on the labeled weights $\boldsymbol{\hat{w}}^{move}$ and $\boldsymbol{\hat{w}}^{mani}$. $\boldsymbol{n}^{move}_{i}$ and $\boldsymbol{n}^{mani}_{i}$ represent the predicted noise values for the arm movement model and the gripper manipulation model, respectively, at the $i$-th denoising step for the noised action sequence. $\boldsymbol{\hat{n}}^{move}$ and $\boldsymbol{\hat{n}}^{mani}$ correspond to the ground truth noise values of the diffusion action models for movement and manipulation, respectively, and are randomly generated.
The inputs to the two action models include the noised actions $\boldsymbol{a}^{move}$ and $\boldsymbol{a}^{mani}$. $\boldsymbol{a}^{move}$ is calculated using the ground truth noise $\boldsymbol{\hat{n}}^{move}$ and the ground truth action $\boldsymbol{\hat{a}}$, while $\boldsymbol{a}^{mani}$ is computed using the ground truth noise $\boldsymbol{\hat{n}}^{mani}$ and the ground truth action $\boldsymbol{\hat{a}}$. The total loss is computed as a weighted sum of the movement loss, the manipulation loss, and the classification loss. The corresponding hyperparameters $w_1$, $w_2$, and $w_3$ are set to $1.0$, $1.0$, and $0.0001$, respectively.
\begin{equation}\label{equ:total_loss}
\begin{aligned}
 L = w_1 * L_{move} + w_2 * L_{mani} +  w_3 * L_{class}.
\end{aligned}
\end{equation}

In the testing phase, the dynamic action model selects and executes the appropriate action model based on the predicted weight $w$. If $w < 0.5$, the model runs the arm movement model, using the high-level VLM-extracted cognition latent $\boldsymbol{f}^{c}$ and the global class token $\boldsymbol{f}^{cls}$ as conditions to predict a sequence of multi-step actions. Otherwise, if $w \geq 0.5$, the model runs the gripper manipulation model, using the cognition latent $\boldsymbol{f}^{c}$ and the local register token $\boldsymbol{f}^{reg}$. Additionally, both models receive random noise $\boldsymbol{n}^{rand}$ as input to facilitate the diffusion process.
\subsection{Dual-Scale Action Weighting}\label{sec:action_weighting}

\begin{figure}[]
    \centering
    \includegraphics[width=1.0\linewidth]{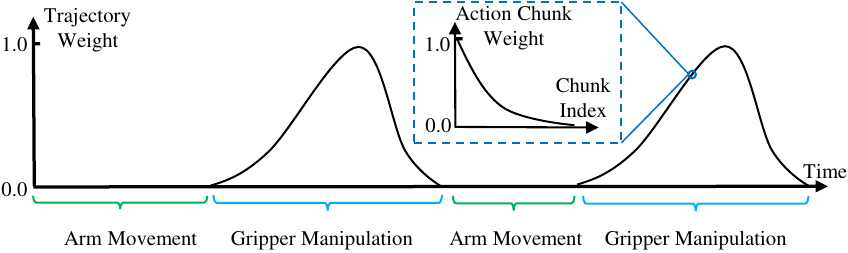}
    \captionsetup{justification=justified, singlelinecheck=false}
    \caption{Illustration of the dual-scale action weighting mechanism. The trajectory weight highlights critical manipulation phases via asymmetrical Gaussian distributions. Within each predicted chunk, the action chunk weight applies exponential decay to prioritize immediate temporal accuracy. The final weight integrates both scales to guide model supervision.
    }
    \label{motivitation}
    \vspace{-1.em}
\end{figure}

To enhance the robustness in distinguishing between arm movement and gripper manipulation, we propose a dual-scale action weighting mechanism for model training, as illustrated in Figure 4. The core objective of the dual-scale action weighting mechanism is to adaptively supervise the learning process by modulating the importance of different action types from both global (trajectory-level) and local (action-chunk-level) perspectives.

Trajectory-level Weights ($\boldsymbol{w}^{t}$): This global perspective segments the entire task trajectory into distinct phases of arm movement and gripper manipulation based on the binary state changes of the robot gripper. For each gripper manipulation process $k$, we employ an asymmetrical Gaussian distribution $\{\mathcal{N}(u,\sigma _{l}^{2}),\mathcal{N}(u,\sigma _{r}^{2})\}$ to define the weights $\boldsymbol{w}_{k}^{t}$. Both distributions share the same mean $u$, representing the temporal midpoint where the gripper state transitions (e.g., from open to closed). To reflect the prior that the action model requires higher precision and supervision focus immediately before the state change, we assign a larger variance to the leading edge ($\sigma _{l}=6$) and a smaller variance to the trailing edge ($\sigma _{r}=2$). The aggregated trajectory weights are defined as $\boldsymbol{w}^{t}=\text{Norm}(\sum _{k}\boldsymbol{w}_{k}^{t})$, representing the normalized sum of all manipulation-related weights. 

Action-chunk-level Weights ($\boldsymbol{w}^{a}$): From a local perspective, we account for the inherent temporal uncertainty in action sequences. Given that prediction confidence typically decays as the temporal distance from the current state increases, we apply an exponentially decreasing function: $\boldsymbol{w}_{j}^{a}=\gamma ^{j}$, where $j$ denotes the index within the action chunk and $\gamma =0.8$ is the decay factor. 

Multi-scale Integration: The final weights are formulated as $\boldsymbol{w}^{move}=(1-\boldsymbol{w}^{t})\odot \boldsymbol{w}^{a}$ and $\boldsymbol{w}^{mani}=\boldsymbol{w}^{t}\odot \boldsymbol{w}^{a}$. By applying these weights to $L_{move}$ and $L_{mani}$, our proposed dual-scale action weighting mechanism dynamically coordinates the arm-movement and gripper-manipulation models through pointwise modulation.

Furthermore, the labeled weight $\hat{w}$ is derived from the trajectory weights: $\hat{w}=1$ if $\boldsymbol{\hat{w}}^{t}>0.5$, and $\hat{w}=0$ otherwise. This weight $\hat{w}$ acts as the ground-truth label for the predicted weight $w$, supervised via the cross-entropy loss $L_{class}$.

\section{Experiments}
\label{sec:Experiments}

We conduct extensive experiments to comprehensively evaluate the performance of our proposed method and to clearly demonstrate its effectiveness in both task-specific and general-purpose manipulation scenarios. 
Specifically, Section~\ref{sec:training} details the training and fine-tuning procedures.
We then present the experimental results on SIMPLER \cite{li2024evaluating} and FurnitureBench \cite{heo2023furniturebench} in Section~\ref{sec:simulation}.
We also conduct real-world evaluations based on a pick-and-place task and present the results in Section~\ref{sec:realworld}.
Section~\ref{sec:ablation} provides an ablation study to analyze the contribution of each component in our framework.

\subsection{Training and Fine-tuning Details}\label{sec:training}
The large-scale Open X-Embodiment Dataset \cite{o2024open} contains over 1 million robot manipulation trajectories collected from across 22 distinct robotic embodiments. We primarily utilize two of its major subsets, Fractal \cite{brohan2022rt} and BridgeDataV2 \cite{walke2023bridgedata}, as our primary pre-training dataset. Our VLA model is trained using a constant learning rate of $2 \times 10^{-5}$ and a batch size of 256 on 8 NVIDIA H100 GPUs for approximately two days.

Furthermore, we fine-tune our DAM-VLA model on both simulated and real-world datasets. For the simulation experiments, we specifically use the FurnitureBench benchmark, which involves contact-rich and long-horizon manipulation tasks. Specifically, we fine-tune our DAM-VLA model on 500 expert trajectories from the ``One-Leg'' assembly task. For real-world evaluation, we construct a pick-and-place scenario in which a Franka robot is teleoperated to pick up a cup and place it into a bowl. A total of 50 demonstrated trajectories were collected for fine-tuning. The fine-tuning process adopts the same hyperparameters as pre-training: a learning rate of $2 \times 10^{-5}$ and a batch size of 256, utilizing 8 NVIDIA H100 GPUs. For a fair comparison, we also fine-tune the OpenVLA and CogACT baselines using the identical datasets.

\begin{table}[]
\setlength{\abovecaptionskip}{0.2cm}
\setlength{\belowcaptionskip}{-0.4cm}
    \begin{center}
    \resizebox{1.0\columnwidth}{!}
    {
    \begin{tabular}{llccccc}\hline
    %Google Robot & 
    \multirow{2}{*}{Method / Google(VA)} & \multicolumn{4}{c}{Success Rates on Different Tasks} & \multirow{2}{*}{Avg} \\
    & PCC & MN & OCD & ODPA &  \\ \hline
    %\multirow{9}{*}{\makecell{Manipulation w. \\ Variant Aggregation \\Setting}}  
    RT-1 \cite{brohan2023rt} &90\%	&46\%	&35\%	&3\%	&44\%  \\
    RT-1-X \cite{vuong2023open} &49\%	&33\%	&29\%	&10\%	&30\%  \\
    RT-2-X \cite{vuong2023open} &82\%   &79\%   &35\%   &21\%   &54\% \\
    Octo-Base \cite{team2024octo} &1\%	&4\%	&1\%	&0\%	&1\%  \\
    RoboVLM \cite{li2024towards}   &76\% &60\% &11\% &- &-\\
    SpatialVLA \cite{qu2025spatialvla} &88\% &73\% &42\% &- &-\\
    OpenVLA \cite{kim2024openvla} &64\%	&64\%	&19\%	&1\%	&37\%  \\
    CogACT \cite{li2024cogact} &96\%  &\textbf{84\%}	&29\%	&40\%	&62\%  \\
    DAM-VLA(Ours) &\textbf{98\%}	&74\%	&\textbf{68\%}	&\textbf{84\%}	&\textbf{81\%}  \\ \hline
    \end{tabular}
    }
    \captionsetup{justification=justified, singlelinecheck=false}
    \caption{Comparison of success rates between our method and existing VLA methods on the Google robot in Variant Aggregation (VA) setting of the SIMPLER simulated evaluation across four tasks.
    (PCC: Pick Coke Can, MN: Move Near, OCD: Open/Close Drawer, ODPA: Open Drawer and Place Apple, Avg: Average)
    %$^+$ refer to \href{https://github.com/allenzren/open-pi-zero?tab=readme-ov-file}{open-pi-zero}.
    }
    \label{Google_robot_VA}
    \end{center}
    % \vspace{-0.6em}
    \end{table}
% VM Table
    
\begin{table}[]
\setlength{\abovecaptionskip}{0.2cm}
\setlength{\belowcaptionskip}{-0.8cm}
    \begin{center}
    \resizebox{1.0\columnwidth}{!}
    {
    \begin{tabular}{llccccc}\hline
    %Google Robot & 
    \multirow{2}{*}{Method / Google(VM)} & \multicolumn{4}{c}{Success Rates on Different Tasks} & \multirow{2}{*}{Avg} \\
    & PCC & MN & OCD & ODPA &  \\ \hline
    %\multirow{9}{*}{\makecell{Manipulation w. \\ Variant Aggregation \\Setting}}  
    RT-1 \cite{brohan2023rt} &87\%	&39\%	&72\%	&8\%	&51\% \\
    RT-1-X \cite{vuong2023open} &59\%	&33\%	&56\%	&17\%	&41\% \\
    RT-2-X \cite{vuong2023open} &79\%  &78\%   &25\%   &4\%   &46\% \\
    Octo-Base \cite{team2024octo} &18\%	&4\%	&24\%	&0\%	&11\% \\
    OpenVLA \cite{kim2024openvla} &14\%	&51\%	&48\%	&0\%	&28\% \\
    RoboVLM \cite{li2024towards}   &77\% &62\% &44\% &- &-\\
    SpatialVLA \cite{qu2025spatialvla} &86\% &78\% &57\% &- &-\\
    CogACT \cite{li2024cogact} &92\%	&82\%	&\textbf{75\%}	&39\%	&72\% \\
    $\pi^{*}_{0}$ %(BF16 Uniform)
    \cite{black2410pi0}$^+$ &89\%	&81\%	&55\%	&53\%	&70\% \\
    DAM-VLA(Ours) &\textbf{96\%}	&\textbf{84\%}	&\textbf{75\%}	&\textbf{78\%}	&\textbf{83\%} \\ \hline
    \end{tabular}
    }
    \captionsetup{justification=justified, singlelinecheck=false}
    \caption{Comparison of success rates between our method and existing VLA methods on the Google robot in Visual Matching (VM) setting of the SIMPLER simulated evaluation across four tasks. 
    (PCC: Pick Coke Can, MN: Move Near, OCD: Open/Close Drawer; ODPA: Open Drawer and Place Apple, Avg: Average; 
    $^+$: \href{https://github.com/allenzren/open-pi-zero?tab=readme-ov-file}{open-pi-zero}, which is trained on Fractal dataset.)
    }
    \label{Google_robot_VM}
    \end{center}
    \vspace{-0.6em}
\end{table}

\begin{table}[]
\setlength{\abovecaptionskip}{0.2cm}
\setlength{\belowcaptionskip}{-0.4cm}
    \begin{center}
    \resizebox{1.0\columnwidth}{!}
    {
    \begin{tabular}{lccccc}\hline
    %WidowX Robot & 
    %Method &\makecell{Put Spoon\\ on Towel} & \makecell{Put Carrot \\on Plater} & \makecell{Stack Green Block \\on Yellow Block} & \makecell{Put Eggplant in \\Yellow Basket} & Avg \\ \hline
    \multirow{2}{*}{Method / WidowX(VM)} & \multicolumn{4}{c}{Success Rates on Different Tasks} &\multirow{2}{*}{Avg} \\
     & SoT & CoP & GoY & EiB & \\ \hline
    %\multirow{10}{*}{\makecell{Manipulation w.\\ Visual Matching \\Setting}}  
    %&
    RT-1-X \cite{vuong2023open}	&4\%	&8\%	&0\%	&0\%	&3\% \\
    Octo-Base \cite{team2024octo}	&7\%	&13\%	&0\%	&44\%	&16\% \\
    Octo-Small \cite{team2024octo}	&47\%	&8\%	&1\%	&51\%	&27\% \\
    OpenVLA \cite{kim2024openvla}	&4\%	&0\%	&0\%	&4\%	&2\% \\
    ECOT \cite{zawalski2024robotic}	&4\%	&8\%	&0\%	&0\%	&3\% \\
    RoboVLM \cite{li2024towards}   &29\% &25\% &13\% &58\% &31\%\\
    SpatialVLA \cite{qu2025spatialvla} &17\% &25\% &\textbf{29\%} &\textbf{100\%} &43\%\\
    CogACT\cite{li2024cogact} &63\%	&50\%	&25\%	&71\%	&52\% \\
    $\pi^{*}_{0}$%(BF16 Uniform)
    \cite{black2410pi0}$^+$ &62\%	&59\%	&24\%	&81\%	&57\% \\
    DAM-VLA(Ours)	&\textbf{88\%}	&\textbf{71\%}	&25\%	&\textbf{100\%}	&\textbf{71\%} \\  \hline
    \end{tabular}
    }
    \captionsetup{justification=justified, singlelinecheck=false}
    \caption{Comparison of success rates between our method and existing VLA methods on the WidowX robot in the Visual Matching (VM) setting of the SIMPLER simulated evaluation across four tasks. 
    (SoT: Put Spoon on Towel, CoP: Put Carrot on Plater; GoY: Stack Green Block on Yellow Block; EiB: Put Eggplant in Yellow Basket; Avg: Average; 
    $^+$: \href{https://github.com/allenzren/open-pi-zero?tab=readme-ov-file}{open-pi-zero}, which is trained on BridgeDataV2.)}
    \label{WidowX_robot}
    \end{center}
    \vspace{-1.5em}
    \end{table}

\subsection{Simulated Evaluations}\label{sec:simulation}
We first evaluate our method using the SIMPLER simulation \cite{li2024evaluating}, a suite of open-source simulated environments designed to mirror common real-world robot manipulation setups. Compared to real-world evaluations, this real-to-sim approach offers a scalable, reproducible, and informative tool that complements high-quality real-world assessments. Extensive testing of various VLA models has shown a strong correlation between evaluations in SIMPLER's simulated environments and real-world performance.
SIMPLER supports two robot embodiments: the Google robot and the WidowX robot. For the Google robot, evaluations are conducted under both Visual Matching (VM) and Variant Aggregation (VA) settings across four tasks, whereas the WidowX robot is evaluated only under the VM setting. The VM setting closely replicates real-world tasks by minimizing discrepancies between simulated and real-world environments. In contrast, the VA setting extends the VM setting by introducing variations in factors such as background, lighting, distractors, and table texture. The success rate of task completion is used as the evaluation metric for all VLA models. Notably, we follow CogACT \cite{li2024cogact} in determining the number of trials conducted in SIMPLER.

Tables~\ref{Google_robot_VA} and \ref{Google_robot_VM} compare our method against existing VLA approaches on the Google robot across four tasks. Our model leads with average success rates of 83\% (VM) and 81\% (VA). Notably, we see substantial gains in the "Open Drawer and Place Apple" task, which requires task-specific manipulation. Moreover, in the VA setting, our success rate markedly exceeds competitors, demonstrating DAM-VLA mitigates performance degradation in dynamic environments.

Table~\ref{WidowX_robot} reports the success rates of our method compared with existing VLA approaches on the WidowX robot in SIMPLER across four tasks under the VM setting. Our model achieves the highest average success rate of 71\%, outperforming competing methods by a substantial margin. In particular, DAM-VLA shows notable improvements in the ``Put Spoon on Towel'' and ``Put Carrot on Plate'' tasks, which involve more diverse object pose variations.

\begin{figure}[]
    \centering
    \includegraphics[width=0.95\linewidth]{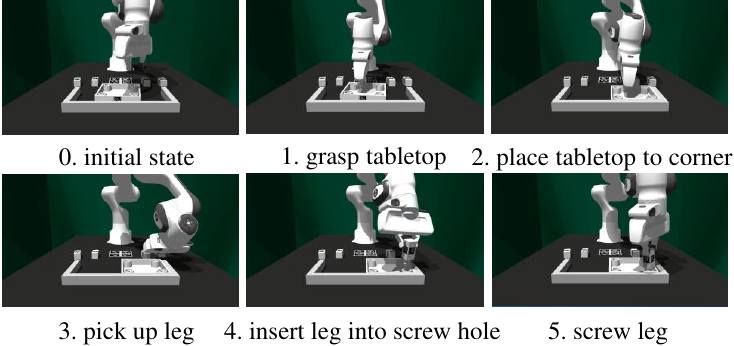}
    \captionsetup{justification=justified, singlelinecheck=false}
    \caption{The entire process of the ``One-Leg'' assembly task in the FurnitureBench environment.}
    \label{distribution}
    % \vspace{-0.2em}
\end{figure}

In addition to the SIMPLER, we further evaluate our method on the ``One-Leg'' assembly task from FurnitureBench \cite{heo2023furniturebench}, which involves contact-rich and long-horizon manipulation. The full task sequence is illustrated in Figure 5. 
We conduct 50 evaluation trials with randomized initial furniture placements. 
As shown in Table \ref{FurnitureBench} are the success rates of each step of the ``One-Leg'' assembly task compared with OpenVLA and CogACT models.
Our method consistently outperforms prior VLA approaches on this challenging furniture assembly task, demonstrating superior generalization and precision under contact-rich manipulation settings. 
More specifically, we achieve a 100\% success rate in the first three steps of ``grasp tabletop'', ``place tabletop to corner'', and ``pick up leg'', and we achieve higher success rates in the contact-rich manipulation ``screw leg''.

% FurnitureBench Table
\begin{table}[]
\setlength{\abovecaptionskip}{0.2cm}
\setlength{\belowcaptionskip}{-0.4cm}
    \begin{center}
    \resizebox{1.0\columnwidth}{!}
    {
    \begin{tabular}{llccccc}\hline
    %FurnitureBench & 
    \multirow{2}{*}{Method / FurnitureBench} & \multicolumn{5}{c}{Success Rates at Each Step} \\
    & 1 & 2 & 3 & 4 & 5 \\ \hline
    OpenVLA \cite{kim2024openvla} &96\%	&94\%	&78\%	&53\%	&29\%  \\
    CogACT \cite{li2024cogact} &98\%  &96\%	&96\%	&56\%	&42\%  \\
    DAM-VLA(Ours) &\textbf{100\%}	&\textbf{100\%}	&\textbf{100\%}	&\textbf{62\%}	&\textbf{56\%}  \\ \hline
    \end{tabular}
    }
    \captionsetup{justification=justified, singlelinecheck=false}
    \caption{The success rate of each step of the ``One-Leg'' assembly task in FurnitureBench is compared with the existing OpenVLA and CogACT methods. The fifth step represents the final success rate of the task.}
    \label{FurnitureBench}
    \end{center}
    \vspace{-1.5em}
    \end{table}
% FurnitureBench Table

\subsection{Real-world Evaluations}\label{sec:realworld}

\begin{figure}[h]
    \centering
    \includegraphics[width=0.95\linewidth]{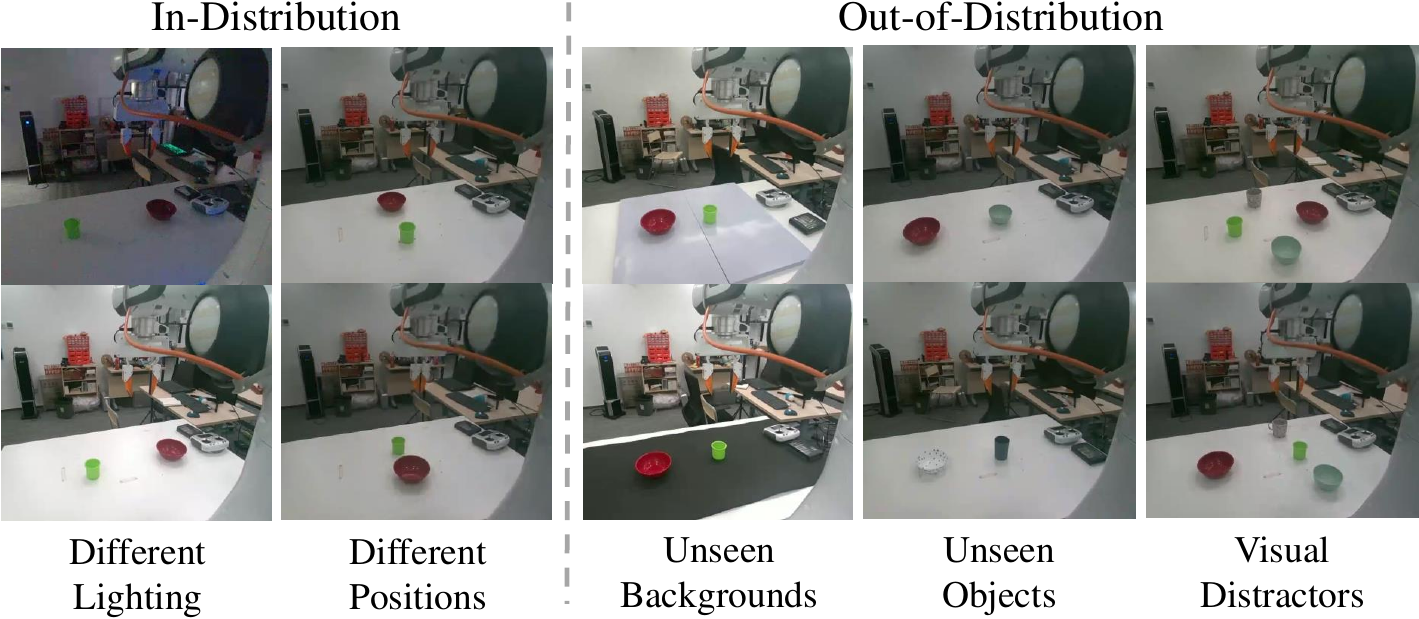}
    \captionsetup{justification=justified, singlelinecheck=false}
    \caption{The evaluation encompasses both in-distribution and out-of-distribution scenarios. The in-distribution setting includes variations in object positions and lighting conditions consistent with the training data, while the out-of-distribution setting introduces novel backgrounds, unseen objects, and visual distractors absent during training.}
    \label{distribution}
    \vspace{-1.0em}
\end{figure}

Our real-world dataset is collected under diverse object placements and lighting conditions. To assess robustness, we divide the evaluation into in-distribution and out-of-distribution scenarios, as illustrated in Figure 6. The in-distribution scenario includes variations in object positions and lighting that are consistent with the training distribution. In contrast, the out-of-distribution scenario introduces previously unseen backgrounds, novel objects, and visual distractors that are absent during training.

Table~\ref{RealDataset} reports the success rates of DAM-VLA compared with the baseline CogACT on the real-world pick-and-place task. To ensure a fair comparison, each method is evaluated with 5 trials per condition across 16 distinct conditions, covering both in-distribution and out-of-distribution scenarios, for a total of 80 trials. The results demonstrate that DAM-VLA consistently outperforms CogACT, achieving higher success rates in both evaluation settings.

\begin{table}[t]
    \begin{center}
    %  \resizebox{0.8\columnwidth}{!}
    % {
    \begin{tabular}{lccc}\hline
    \multirow{2}{*}{Method / Real-world} & \multicolumn{3}{c}{Success Rates on Different Scenarios} \\
    & ID	& OOD	&Avg \\ \hline
    CogACT	&65.7\%	&60.0\%	&62.9\%  \\ 
    DAM-VLA (Ours)	&\textbf{91.4\%}	&\textbf{82.2\%}	&\textbf{86.8\%}  \\ \hline
    \end{tabular}
    % }
    \caption{The success rates of our method compared with CogACT on the pick-and-place task in the real-world evaluation. ID: In-Distribution, OOD: Out-of-Distribution.}
    \label{RealDataset}
    \end{center}
    \vspace{-1.0em}
    \end{table}

\subsection{Ablation Study}\label{sec:ablation}

\begin{table}[]
\setlength{\abovecaptionskip}{0.2cm}
\setlength{\belowcaptionskip}{-0.2cm}
    \begin{center}
    \resizebox{1.0\columnwidth}{!}
    {
    \begin{tabular}{ccccccccc}\hline
    \multicolumn{5}{c}{Components} &\multicolumn{4}{c}{Success Rates} \\
    VT &ACW &TW &DAM &DL &\makecell{Google\\(VM)} &\makecell{Google\\(VA)} &\makecell{WidowX\\(VM)} &Avg  \\ \hline  
-	&-	&-	&-	&-	&64\%	&61\%	&50\%	&58\%  \\
\checkmark	&\checkmark	&-	&-	&-	&78\%	&68\%	&53\%	&66\%  \\
\checkmark	&\checkmark	&\checkmark	&-	&-	&76\%	&63\%	&51\%	&63\%  \\
\checkmark	&\checkmark	&\checkmark	&\checkmark	&-	&82\%	&72\%	&43\%	&66\%  \\
\checkmark	&\checkmark	&\checkmark	&\checkmark	&\checkmark	&83\%	&\textbf{81\%}	&\textbf{71\%}	&\textbf{78\%}  \\ 
-	&\checkmark	&\checkmark	&\checkmark	&\checkmark	&\textbf{84\%}	&75\%	&58\%	&73\%  \\ 
-	&-	&-	&\checkmark	&\checkmark	&66\%	&64\%	&49\%	&60\%  \\ \hline  
    \end{tabular}
    }
    \captionsetup{justification=justified, singlelinecheck=false}
    \caption{An ablation study is conducted on the WidowX robot in Visual Matching (VM) setting and the Google robot in both VM and Variant Aggregation (VA) settings of the SIMPLER environment.}
    \label{Ablation_study}
    \end{center}
    \vspace{-1.5em}
    \end{table}

%In Table \ref{Ablation_study}, 
For the ablation studies, we employ the SIMPLER evaluation environment using the WidowX robot in the VM setting and the Google robot in both the VM and VA settings. To assess the contribution of different components of our method, we conduct a detailed analysis of their corresponding success rates. As shown in Table~\ref{Ablation_study}, the main components under investigation include:
(1) \textbf{Visual Tokens (VT)}: the visual class and register tokens output by the vision model, used by the action models to predict actions.
(2) \textbf{Action Chunk Weights (ACW)}: the action-chunk weights used to supervise DAM-VLA.
(3) \textbf{Trajectory Weights (TW)}: the trajectory weights used to supervise DAM-VLA.
(4) \textbf{Dynamic Action Model (DAM)}: consisting of the arm movement model and the gripper manipulation model.
(5) \textbf{Dual Latents (DL)}: the reasoning latent and cognition latent extracted from different transformer layers of the LLM.
Starting from a baseline VLA model with a single action model, adding VT and ACW improves performance on both the Google and WidowX robots. Since current VLA consists of only a single diffusion action model, VT here only contains the class token. Incorporating TW, which is specifically designed for dual action models, leads to a slight decrease in performance on both robots, which is expected. Adding DAM and DL subsequently yields the highest average success rate of 78\%, which demonstrates the effectiveness of our DAM-VLA framework. When the VLA is configured with dual action models, VT contains both the class and register tokens. Finally, removing VT and the dual-scale action weighting mechanism (ACW and TW) from DAM-VLA results in a substantial performance drop, which further emphasizes the critical role of both the dynamic action model and the dual-scale action weighting mechanism. Notably, removing VT means removing both the class and register tokens.

%===============================================================================

\section{Conclusion}
\label{sec:conclusion}
Our proposed DAM-VLA method dynamically integrates the inherent reasoning capabilities of VLMs with specialized diffusion-based action models designed for arm movement and gripper manipulation. Extensive experiments demonstrate that our approach not only significantly outperforms existing VLA methods but also delivers stable results in both task-specific and general manipulation scenarios within dynamic environments. By dynamically routing between specialized action models based on VLM-guided cues, DAM-VLA paves a new path for incorporating semantic understanding into embodied decision-making. Its demonstrated generalization ability and stability across diverse tasks and dynamic settings highlight its potential as a foundational framework for next-generation adaptable robotic systems in real-world applications.

While DAM-VLA achieves strong performance, several aspects remain open for further improvement. First, although our experimental evaluation already involves contact-rich and long-horizon tasks, it is currently centered on pick-and-place and furniture assembly; expanding to more diverse task families will provide broader validation. Second, while DAM-VLA substantially improves the success rate of most tasks, the relatively modest gains in the ``Stack Green Block on Yellow Block'' scenario indicate opportunities to enhance fine-grained coordination and stability. Finally, DAM-VLA currently routes between only two action models; generalizing this mechanism to handle multiple action types with richer routing signals will further broaden its applicability to complex real-world settings.

\bibliographystyle{IEEEtran}
\bibliography{root}

\end{document}